# Free-form Grid Structure Form Finding based on Machine Learning and Multi-objective Optimisation

Yiping MENG*, Yiming SUN[a]

*School of Computing, Engineering & Digital Technologies, Teesside University
Tees Valley, Middlesbrough TS1 3BX
y.meng@tees.ac.uk

[a] Department of Automatic Control and System Engineering, University of Sheffield

**Abstract**

Free-form structural forms are widely used to design spatial structures for their irregular spatial morphology. Current free-form form-finding methods cannot adequately meet the material properties, structural requirements or construction conditions, which brings the deviation between the initial 3D geometric design model and the constructed free-form structure. Thus, the main focus of this paper is to improve the rationality of free-form morphology considering multiple objectives in line with the characteristics and constraints of material. In this paper, glued laminated timber is selected as a case. Firstly, machine learning is adopted based on the predictive capability. By selecting a free-form timber grid structure and following the principles of NURBS, the free-form structure is simplified into free-form curves. The transformer is selected to train and predict the curvatures of the curves considering the material characteristics. After predicting the curvatures, the curves are transformed into vectors consisting of control points, weights, and knot vectors. To ensure the constructability and robustness of the structure, minimising the mass of the structure, stress and strain energy are the optimisation objectives. Two parameters (weight and the z-coordinate of the control points) of the free-from morphology are extracted as the variables of the free-form morphology to conduct the optimisation. The evaluation algorithm was selected as the optimal tool due to its capability to optimise multiple parameters. While optimising the two variables, the mechanical performance evaluation indexes such as the maximum displacement in the z-direction are demonstrated in the 60th step. The optimisation results for structure mass, stress and strain energy after 60 steps show the tendency of oscillation convergence, which indicates the efficiency of the proposal multi-objective optimisation.

## 1. Introduction

As the regular architecture form could not meet the up-to-date aesthetic requirements, the distinct architectural expression and engineering challenges of free-form grid structures place them at the forefront of spatial design innovation [1]. The irregular forms put forward new challenges for the conventional design paradigms from design methods to design tools. These structures defy traditional modelling with complex surfaces that cannot be succinctly expressed through simple analytic functions, departing from conventional architectural forms. Historically, the genesis of such structures relied heavily on physical experimentation methods, such as the inverse hanging method for compression structures and the soap film technique for pre-stressed shapes [2], [3]. The emergence of computational graphics technologies, notably Bezier surfaces, B-splines, and Non-Uniform Rational B-splines (NURBS), has revolutionised the field of complex geometry design and optimisation [4], [5].

To achieve the complex form, some new criteria need to be met compared with traditional regular architectural design, such as smoothness, and geometric dimensions [6]. In addition, the deviation of the real project from the original design needs to be decreased if considering constraints like material, structural stiffness and manufacturing, which means that additional knowledge about geometry is essential in the optimum design [7].

Traditional design and optimization approaches often fall short in addressing these complex requirements, spotlighting the necessity for sophisticated optimization techniques capable of untangling the complexities of free-form structure [8]. The demand for new design techniques has led to the adoption and development of diverse computational methods and algorithms. Among these new techniques, optimisation algorithms [9] stand out due to their efficacy in addressing the multifaceted challenges inherent in the design and realisation of optimised structures. Related studies have shown that optimisation for structural performance effectively generates free-from structures with stable mechanical behaviour [10]. Different algorithms have been applied and developed to the topology and shape optimisation [11], for example, gradient descent [12], GA (Genetic Algorithm) [13], evolution algorithm [14] and NGSA-II algorithm [15]. Among the different algorithms, the optimisation objectives are various, including maximum displacement, element stress overall quality of the structure, and strain energy [16] by setting the coordinate of control points as the variable. However, in these multi-objective optimisation processes, the efficiency of the existing algorithm is not high, and the optimal results have low similarity to the initial surface before the optimisation [14].

Under the context of calling for new technologies for rational free-form structures, machine learning (ML) has also shown its potential to generate free-form structures based on its data processing capability [17]. ML is a new ground-breaking technique that has been widely used in computer vision, image processes, natural language processes, and generative tasks [18]. In [19], augmenting finite element analysis for optimizing space frame structures, aiming to reduce computation times significantly through ML has been discussed. Based on the learning and analysis of the data from the collected data (e.g. images, graphics), ML can generate new data of the same type through deep neural networks (DNNs) [20] and generative adversarial networks (GAN) [21] in generative floor plans [22]. Other ML networks like long short-term memory (LSTM) can be applied to dealing with graphical information [23]. Since the complex geometric information of the free-form morphology can not be fully represented through graphics or images, the 2D application of these ML methods is one main limitation.

In this study, ML is utilised to generate rational geometric information for free-form morphology considering the constraints from material properties by taking timber as the building material. Based on the predicted geometric information, the free-form morphology is further optimised for structural performance through evolution algorithms.

## 2. Machine Learning Prediction for Material Rationality

Quantifying the impact of building material properties on free-form morphology is challenging. The primary approach to using ML to predict the geometric information of free-form morphology involves using real free-form grid structures as the learning input. This input encompasses data about the extent to which building materials can be shaped. After training, ML can predict new curves based on the range of achievable shapes, as learned from previous cases of free-form structures. One significant advantage of using ML for prediction is its ability to integrate the design with material considerations effectively.

### 2.1. Geometric Description for Free-form Morphology

The mathematical model serves as a superior approximation method due to its high efficiency and accuracy. Among the various mathematical models, B-splines and NURBS are the most commonly utilized. Based on the rational B-spline, NURBS is developed by adding an extra parameter called weights. NURBS offers greater flexibility and adaptability across diverse geometric types than B-splines. Consequently, NURBS is the preferred method for describing the geometry of free-form structures.

According to the definition of NURBS, the main factors that control the shape of the NURBS surface (or curve) are (1) the location of the control points, the number; (2) the weight factor, generally the larger the weight factor, the closer the surface (or curve) is to the control points; (3) the knot distribution of parameters u,v and the order of the surface. The NURBS curve with control points is shown in Figure 1(a) and (b) shows the relationship between the NURBS surface and control points.

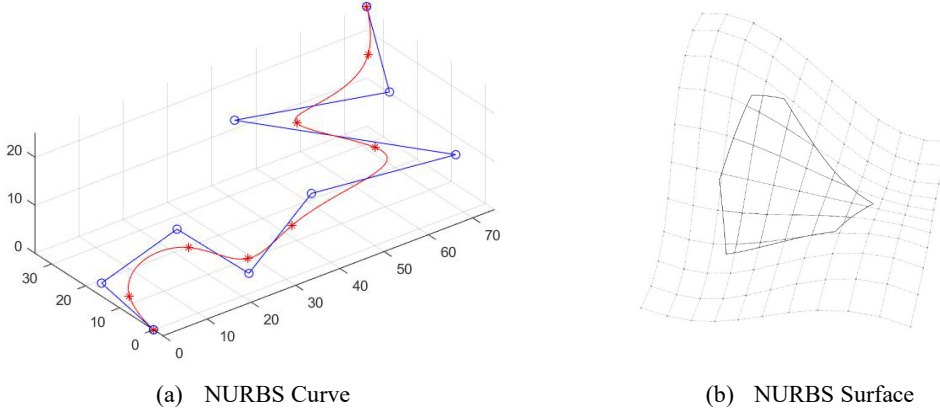

(a) NURBS Curve  (b) NURBS Surface

Figure 1   NURBS curve and surface with control points

A $p$ times NURBS curve is defined as:

$$C(U) = \frac{\sum_{i=0}^{n} N_{i,p}(u)\omega_i P_i}{\sum_{i=0}^{n} N_{i,p}(u)\omega_i}, \quad a \leq u \leq b \tag{1}$$

$C(u)$ means the coordinate of a random point on the NURBS curve in x-y-z space, $\{P_i\}$ is the control point, $N_{i,p}(u)$, $i = 0,1,\dots,n$ is the ith p times B-spline base function, which is called B-spline. $\{\omega_i\}$ is the weight factor. The coordinate of a random point on the NURBS surface can be expressed in the below formulation:

$$S(u,v) = \frac{\sum_{i=0}^{m}\sum_{j=0}^{n} N_{i,p}(u)N_{j,q}(v)\omega_{i,j} P_{i,j}}{\sum_{i=1}^{m}\sum_{j=1}^{n} N_{i,p}(u)N_{j,q}(v)\omega_{i,j}}, a \leq u \leq b; c \leq v \leq d \tag{2}$$

Where: $u,v$ are the parameters of the surface; $p$, $q$ are the number of powers of the surface; the surface is the segmentation functions about u,v; knot vectors $U,V$ are combined by knots $u,v$. For curved surfaces, $\{P_{(i,j)}\}$ forms a control grid in two directions and the number of control points is $(n+1) \times (m+1)$; $N_{i,p}(u)$ and $N_{i,q}(v)$ are base functions of $u$ and $v$ directions; $\{\omega_{(i,j)}\}$ is the weight factor

**2.2. Machine Learning Prediction Experiment**

Selecting the appropriate ML method for prediction tasks is crucial as it determines the types of data used for training and testing. In [24], the free-form structure is depicted through curves, which are then transformed into sequential datasets. Building on this method of information transformation, the Transformer model is chosen for handling the transformed sequential data in this study. Originally designed for natural language processing [25], Transformers have been successfully adapted for various sequence modelling tasks across spatial domains [26]. Their self-attention mechanism provides a sophisticated means to model relationships between different points along a curve, effectively capturing both local and global dependencies.

To complete the prediction learning task, choosing the appropriate free-form case is critical. For this ML process, the Center Pompidou-Metz Model is utilized as a case study to extract data for the Transformer. The structural design of the Centre Pompidou-Metz features a weave pattern, constructed from glue-laminated timber, making it an excellent source of learning input shown in Figure 2.

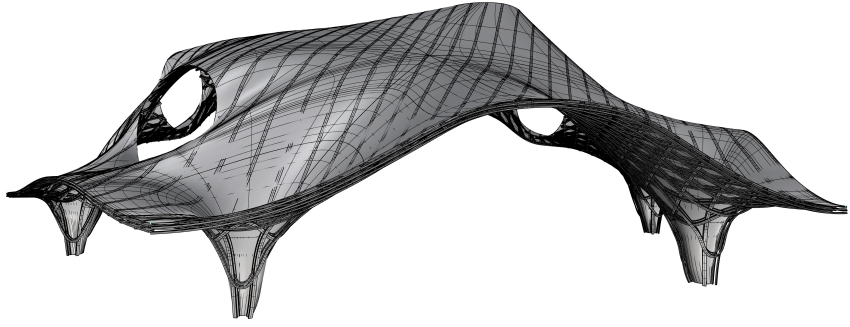

Figure 2 3D model of the Centre Pompidou-Metz case

*2.2.1. Data Transformation*

The main difficulty for ML prediction application in free-form structural morphology is the extraction of the information, as most of the ML deals with the data in 1 or 2 dimensions. In this model, all timber beams and columns are curved to create a distinctive free-form shape. An essential step following the 3D modelling process is to extract geometric information and convert it into discrete numerical values to serve as inputs for ML. Each beam or column in this model is characterised by six faces and 12 boundary lines, which include four critical curves that define the geometry necessary to generate this unique curved structure. The data transformation process is presented in Figure 3.

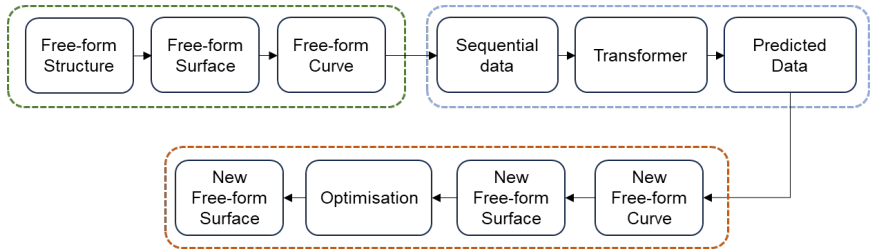

Figure 3 Data extraction and transformation process

*2.2.2. Training and Learning*

After establishing the dataset format, identifying the relevant features and the desired outputs is crucial for the prediction task. In this study, geometric information representing the free-form NURBS curve is extracted, including x-coordinate, y-coordinate, z-coordinate, position parameter, curvature, vector_x, vector_y, and vector_z. The features used to predict the curvature, vector_x, vector_y, and vector_z are the x-coordinate, y-coordinate, z-coordinate, and position parameters. For the pilot test, 16 curves are selected, each divided into 20 segments, resulting in 336 point samples. To enhance the dataset, K-fold validation is adopted, creating five folds with sample sizes of 3, 3, 3, 3, and 4, respectively. Following dataset preparation, the subsequent steps using the Transformer model include:

After preparing the dataset, the following steps of using Transformer are as follows:

- Positional encoding: This allows the model to recognize the position of each point in the sequence, crucial for maintaining the order of data in sequence processing;
- Transformer encoder layer: This layer, which can be stacked, forms the encoder part of the Transformer, essential for processing the sequence data;
- Modify the Transformer model: Adaptations are made so the model can take sequences of curve data as input and predict the desired geometric outputs;
- Train the model: The model is trained with a batch size of 32 and over 100 epochs to ensure adequate learning.

- Evaluation of the model: Performance is assessed on a test set to gauge the effectiveness of the model under evaluation conditions;
- Fine-tuning and Optimisation: Depending on initial results, adjustments are made to the model's architecture, training parameters, and learning rate to optimize performance.

By following the steps, the value of the loss function is shown in Figure 4. The training and testing results are shown in Figure 5.

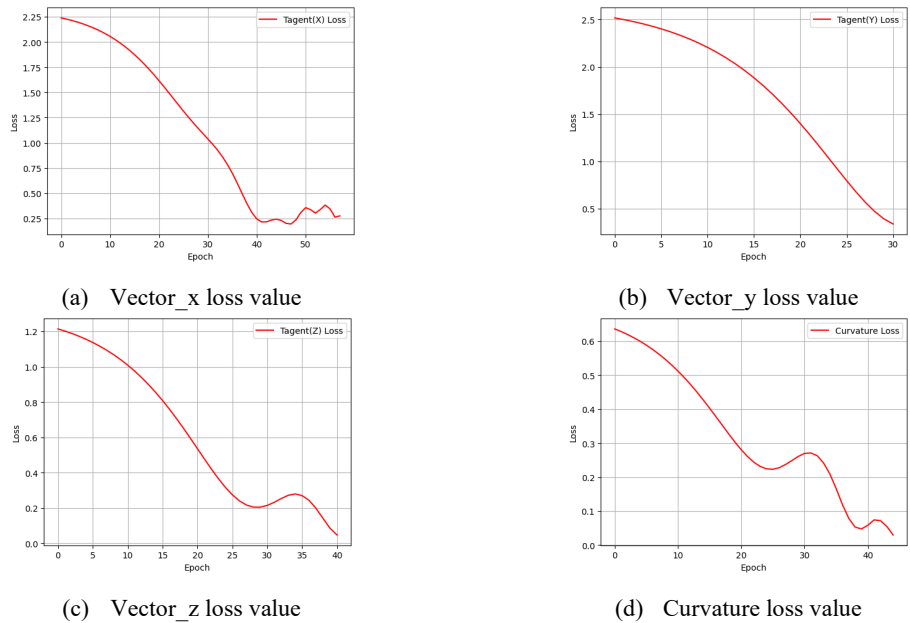

(a) Vector_x loss value  (b) Vector_y loss value

(c) Vector_z loss value  (d) Curvature loss value

Figure 4 Loss function value

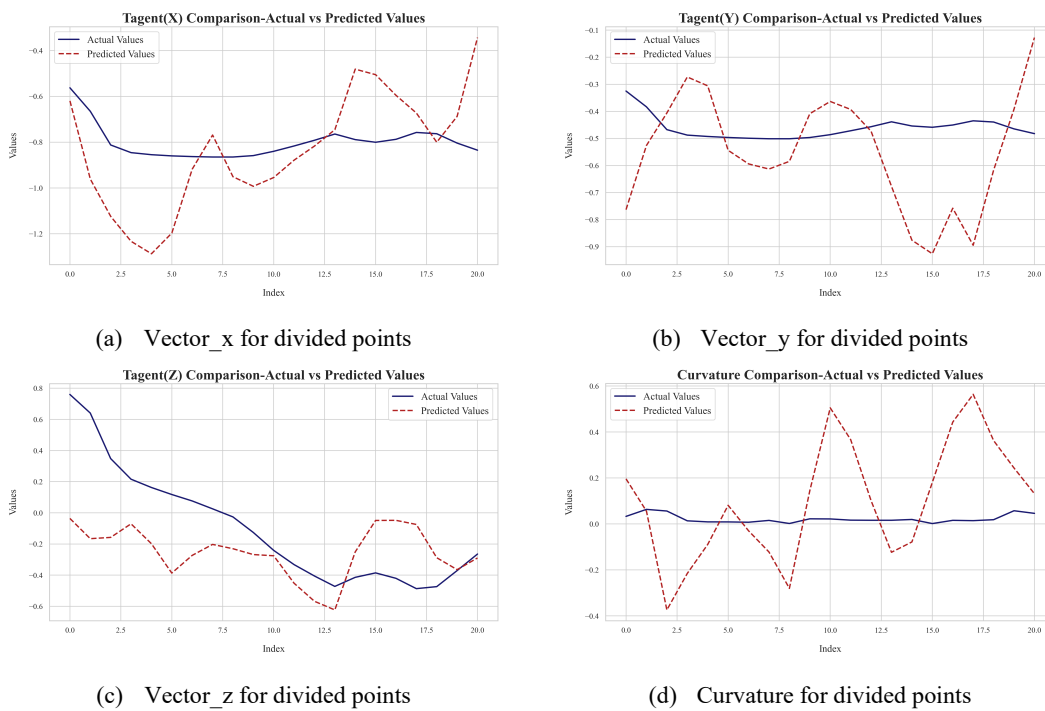

(a) Vector_x for divided points  (b) Vector_y for divided points

(c) Vector_z for divided points  (d) Curvature for divided points

Figure 5 Test results compared with the actual values

The loss function graphs for the curvature and the tangent vectors (X, Y, Z) depict the model's learning curve over successive epochs. The downward trajectory of the loss across all graphs signifies the Transformer model's ability to minimize error and improve its predictions over time. As seen in the provided graphs, there are observable gaps between the predicted and actual values for curvature and the components of the tangent vector (X, Y, and Z). Despite these discrepancies, the overall trend captured by the Transformer indicates a strong alignment with the actual data, substantiating the model's validity in capturing the essential patterns and tendencies of the structure's geometric properties.

## 3. Multi-objective Optimisation for Free-Form Morphology

Based on the tuned Transformer model, given the x-coordinate, y-coordinate, z-coordinate, and position parameter of the points, the NURBS curves meet the restrictions of the GLT and can be interpolated through curvature, vector_x, vector_y, and vector_z. In this case, three curves are selected to patch the free-form prototype surface in Figure 6 (a), which is arrayed to generate the surface shown in Figure 6 (b).

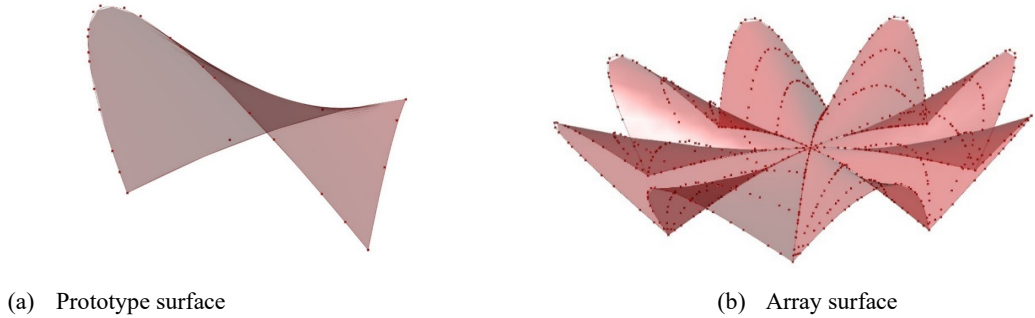

(a)  Prototype surface                                      (b)  Array surface

Figure 6 Example surface

### 3.1. Strain Energy for Structural Rationality

The optimisation method and objectives for structural mechanical rationality have a direct impact on the outcome of the morphology. The optimal objective is commonly set as displacement, stress, strain energy and others, and displacement and stress are vectors that reflect the structure's local characteristics, whereas stain energy is scalar.

The determination of the interrelationships between mechanical properties is a critical issue to address during the morphology creation process. The structural balance equation is:

$$F = K \cdot \delta \qquad (3)$$

F – Force vector of structure nodes

K – Structural stiffness matrix

$\delta$ – Displacement vector of the structure

When the structure is subjected to small elastic deformation, the strain energy $U$ is expressed as:

$$U = \frac{1}{2}F^T\delta \qquad (4)$$

Eq (4) deduces the relationship between strain energy and structure internal force. Research has demonstrated that when strain energy is selected as the objective function, as the strain energy of the structure decreases, not only does the stiffness of the structure increase, but the bending moments are greatly reduced, increasing the ultimate load capacity. Assuming that the load is constant, the structure's strain energy is proportional to the displacement of the structural nodes; that is, decreasing the displacement results in decreasing the strain energy. The smallest strain energy, smallest structural displacement, and largest structural rigidity are all mutually unified.

The strain energy is a scalar, and its value can be thought of as the sum of the strain energies of all the elements in the structure. Besides, the strain energy is unrelated to the selected coordinate system. In the global coordinate system, the strain energy equals the strain energy in the local coordinate system. In complex structure systems, the finite element method is commonly used to divide the structure into several elements when calculating them.

$$U = \sum_{i=1}^{N} \bar{u}_i \tag{5}$$

Where:

$\bar{u}_i$ – Strain energy of every element

$N$ – Total number of the elements in the structure

### 3.2. Multi-objective Optimisation for Free-form Morphology

To conduct multi-objective optimisation for the example free-form surface in Figure 6, the optimal model can be expressed as a nonlinear function on multidimensional space formed by vector $P$:

$$min \begin{Bmatrix} U(P) \\ Mass \\ \sigma \end{Bmatrix} \tag{6}$$

The optimised objectives are set as the strain energy of gravity load and 0.02 mesh load, the mass, and the max stress in the z-direction Finding the minimum of $U(P)$ can be seen as solving the extreme value problem for nonlinear multivariate functions. The solution to the extreme value problem can be achieved by a one-dimensional search, which means the direction of descent of the objective function $U(P)$, which is used to adjust the optimisation variables $P$ to gradually converge to the optimal solution.

In the example free-form surface, the three curves are divided into 4 control points and 22 control points, respectively. The optimisation variables are chosen as the z-coordinate of the two control points of the two side curves and the weights of the 11th and 12th control points and the optimal results are shown in Figure 7. The optimal value for the mass remained within a narrow range, indicating a consistent reduction of about 3.6% from the original value. The curve's trend, displaying convergence in oscillation, underscores the effectiveness of the multi-objective optimization process. The optimal stress value achieved was 0.005, marking a substantial decrease from the original value of 0.013. The results fluctuated within a range of 0.005 to 0.011, with a minimum reduction of 15%. While the reduction in strain energy under gravity load—from 0.02 to 0.018—was modest, the results for the mesh load were more dramatic, with a decrease from nearly 3.5 to 1.1, corresponding to a 68% reduction. The visualised optimised results and comparisons of the initial, and 60th steps are compared in Figure 8 and Figure 9.

### 4. Discussion

The Transformer model has demonstrated exceptional effectiveness in handling the complexities inherent in free-form designs. Its capacity to process sequential 3D geometric data enables accurate predictions of structural forms, showcasing its potential beyond traditional natural language processing tasks and extending its utility to spatial and geometric data analysis. Despite some variation between predicted and actual values, this does not significantly detract from the Transformer's utility in predictive tasks. These discrepancies are largely due to the inherent complexities of free-form structures and the challenges in capturing the nuances of three-dimensional geometry through machine learning.

The integration of evolutionary algorithms was pivotal in refining the Transformer-generated designs to meet specific structure performance criteria. Significant improvements were observed in multi-objective optimization concerning mass, stress, and strain energy, validating the multifaceted capabilities of these algorithms. The evolutionary approach effectively enhanced the exploration and exploitation phases of the optimization process, leading to more efficient and robust design solutions.

However, this research faces several notable limitations that could influence the applicability and effectiveness of the proposed methodologies in various free-form structure scenarios. Firstly, the quality and quantity of data are crucial for training the machine learning models, and any inadequacies in this regard can significantly impair model accuracy. Ensuring a comprehensive and high-quality dataset is paramount for achieving reliable predictions and robust optimization results.

Furthermore, the generalizability of the model when applied to structures or materials different from those used in this study is a concern. Adapting the model to accommodate larger or more intricate designs necessitates complex modifications, which may pose challenges in terms of computational resources and the need for tailored adjustments. This highlights the importance of further research into developing more adaptable and scalable models that can handle a broader range of free-form structure designs without compromising performance.

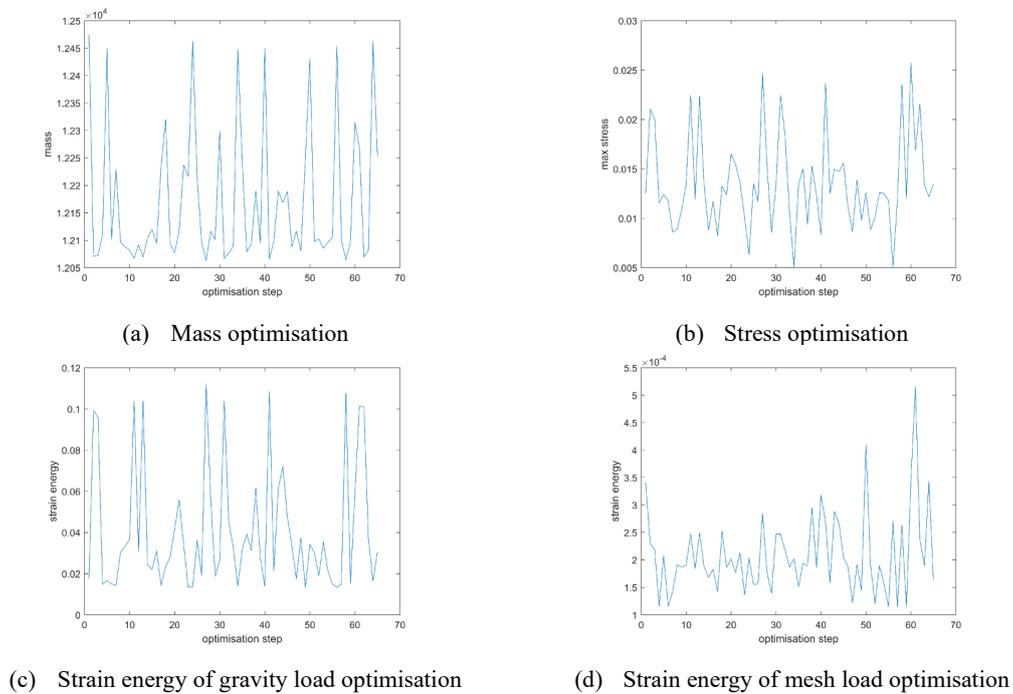

(a) Mass optimisation  (b) Stress optimisation

(c) Strain energy of gravity load optimisation  (d) Strain energy of mesh load optimisation

Figure 7 Optimisation results

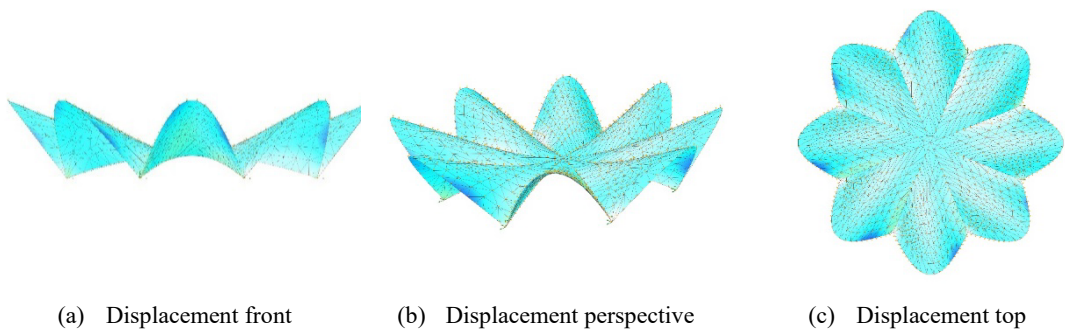

(a) Displacement front  (b) Displacement perspective  (c) Displacement top

Figure 8 Displacement of original morphology

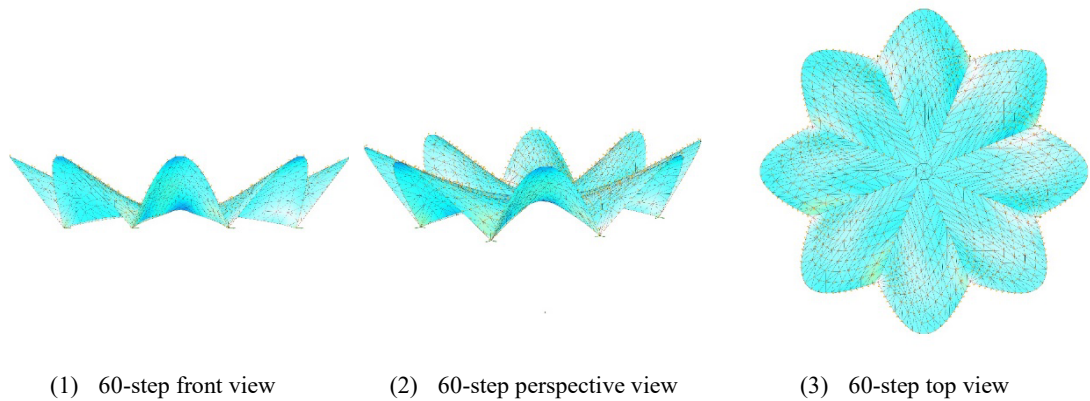

    (1) 60-step front view        (2) 60-step perspective view        (3) 60-step top view

Figure 9 Displacement after 60-step evolution

## 5. Conclusion

This study represents a significant advancement in the field of architectural design by integrating machine learning and evolutionary algorithms to generate and optimise free-form morphology. This approach successfully addresses the dual challenges of adhering to building material limitations and enhancing structural stiffness, which are critical in the construction of viable and sustainable architectural structures. By extracting geometric information from free-form surfaces and translating these into sequential data, the use of advanced machine learning techniques is facilitated, specifically the Transformer model, to predict and optimise structural forms. This approach not only streamlines the initial stages of design but also ensures that the final forms are both feasible and structurally rational. despite some deviations from the actual values, the Transformer's predictive outputs are promising. The model has effectively demonstrated its applicability to geometric form prediction in the context of free-form surface analysis, marking a significant step forward in the intersection of machine learning and architectural design. Still, there are limitations in the application of the methodology. The primary constraint is the reliance on the precision of 3D geometric models as learning inputs, which may not capture the full complexity of real construction projects. Further research is needed to expand the capabilities of the algorithms used, incorporating more dynamic and real-time data inputs involving variations in material properties, construction tolerances, and environmental impacts. The oscillatory convergence patterns of the multi-objective evolutionary optimisation suggest that the stability of the algorithms could be further improved to ensure smoother progression towards the optimum.

## Acknowledgements

This study benefited from the collaborative efforts of all the participants. Special thanks to Dr Yiming Sun from the University of Sheffield for his valuable insight that significantly enhanced the quality of this research. Dr Sun greatly contributed to the modelling and tuning of the Transformer model. Dr Yiping Meng from Teesside University made efforts in geometric modelling, data preparation and optimisation.

## References

[1]	H. Pottmann, A. Schiftner, and J. Wallner, 'Geometry of architectural freeform structures.', in *Symposium on solid and physical modeling*, Citeseer, 2008, p. 9.

[2]	K.-U. Bletzinger and E. Ramm, 'Structural optimization and form finding of light weight structures', *Computers & Structures*, vol. 79, no. 22–25, pp. 2053–2062, 2001.

[3]     F. Otto and B. Rasch, 'Form finding', *Edition Axel Menges. Stuttgart, Germany*, 1995.

[4]     N. Vukašinović and J. Duhovnik, 'Advanced CAD modeling', *Explicit, parametric, free-form CAD and Re-engineering. Cham: Springer Nature Switzerland AG*, 2019.

[5]     H. Ghasemi, R. Brighenti, X. Zhuang, J. Muthu, and T. Rabczuk, 'Optimal fiber content and distribution in fiber-reinforced solids using a reliability and NURBS based sequential optimization approach', *Structural and Multidisciplinary Optimization*, vol. 51, pp. 99–112, 2015.

[6]     H. Pottmann, M. Eigensatz, A. Vaxman, and J. Wallner, 'Architectural geometry', *Computers & graphics*, vol. 47, pp. 145–164, 2015.

[7]     J. Wallner and H. Pottmann, 'Geometric computing for freeform architecture', *Journal of Mathematics in Industry*, vol. 1, pp. 1–19, 2011.

[8]     G. C. Marano, G. Quaranta, and R. Greco, 'Multi-objective optimization by genetic algorithm of structural systems subject to random vibrations', *Structural and multidisciplinary optimization*, vol. 39, pp. 385–399, 2009.

[9]     Q. Xia, M. Y. Wang, S. Wang, and S. Chen, 'Semi-Lagrange method for level-set-based structural topology and shape optimization', *Structural and multidisciplinary optimization*, vol. 31, pp. 419–429, 2006.

[10]    M. Shimoda, T. Okada, T. Nagano, and J.-X. Shi, 'Free-form optimization method for buckling of shell structures under out-of-plane and in-plane shape variations', *Structural and Multidisciplinary Optimization*, vol. 54, pp. 275–288, 2016.

[11]    S. Çarbaş and M. P. Saka, 'Optimum topology design of various geometrically nonlinear latticed domes using improved harmony search method', *Structural and Multidisciplinary Optimization*, vol. 45, pp. 377–399, 2012.

[12]    C. Le, T. Bruns, and D. Tortorelli, 'A gradient-based, parameter-free approach to shape optimization', *Computer Methods in Applied Mechanics and Engineering*, vol. 200, no. 9–12, pp. 985–996, 2011.

[13]    D. E. Goldberg, 'Optimization, and machine learning', *Genetic algorithms in Search*, 1989.

[14]    Z. Wang, Z. Cao, F. Fan, and Y. Sun, 'Shape optimization of free-form grid structures based on the sensitivity hybrid multi-objective evolutionary algorithm', *Journal of Building Engineering*, vol. 44, p. 102538, 2021.

[15]    Q. Ma, M. Ohsaki, Z. Chen, and X. Yan, 'Multi-objective optimization for prestress design of cable-strut structures', *International Journal of Solids and Structures*, vol. 165, pp. 137–147, 2019.

[16]    H. Ohmori, T. Kimura, and A. Maene, 'Computational morphogenesis of free form shells', in *Symposium of the international association for shell and spatial structures (50th. 2009. Valencia). Evolution and trends in design, analysis and construction of shell and spatial structures: Proceedings*, Editorial Universitat Politècnica de València, 2009.

[17]    H. Huang, E. Kalogerakis, and B. Marlin, 'Analysis and synthesis of 3D shape families via deep-learned generative models of surfaces', in *Computer graphics forum*, Wiley Online Library, 2015, pp. 25–38.

[18]    P. P. Shinde and S. Shah, 'A review of machine learning and deep learning applications', in *2018 Fourth international conference on computing communication control and automation (ICCUBEA)*, IEEE, 2018, pp. 1–6.

[19]    Z. Aksöz and C. Preisinger, 'An interactive structural optimization of space frame structures using machine learning', in *Impact: Design with all senses: Proceedings of the design modelling symposium, berlin 2019*, Springer, 2020, pp. 18–31.

[20]    H. Larochelle, Y. Bengio, J. Louradour, and P. Lamblin, 'Exploring strategies for training deep neural networks.', *Journal of machine learning research*, vol. 10, no. 1, 2009.

[21]    I. J. Goodfellow *et al.*, 'Generative adversarial networks. arXiv preprint arXiv: 14062661'. 2014.

[22]    W. Huang and H. Zheng, 'Architectural drawings recognition and generation through machine learning', in *Proceedings of the 38th annual conference of the association for computer aided design in architecture, Mexico City, Mexico*, 2018, pp. 18–20.

[23]    K. Xie and Y. Wen, 'LSTM-MA: A LSTM method with multi-modality and adjacency constraint for brain image segmentation', in *2019 IEEE international conference on image processing (ICIP)*, IEEE, 2019, pp. 240–244.

[24]    Y. Meng, Y. Sun, and W.-S. Chang, 'Morphology of free-form timber structure determination by LSTM oriented by robotic fabrication', in *The international conference on computational design and robotic fabrication*, Springer, 2022, pp. 466–477.

[25]    T. Lin, Y. Wang, X. Liu, and X. Qiu, 'A survey of transformers', *AI open*, vol. 3, pp. 111–132, 2022.

[26]    C. Zhu *et al.*, 'Long-short transformer: Efficient transformers for language and vision', *Advances in neural information processing systems*, vol. 34, pp. 17723–17736, 2021.